\documentclass[runningheads]{llncs}
\usepackage{graphicx}

\usepackage{tikz}
\usepackage{comment}
\usepackage{amsmath,amssymb} 
\usepackage{color}

\usepackage[accsupp]{axessibility}  

\usepackage[pagebackref,breaklinks,colorlinks]{hyperref}

\usepackage{bm}
\usepackage[font=small,labelfont=bf]{caption}
\usepackage{subcaption}
\usepackage{bold-extra}
\usepackage{xspace}
\usepackage{wrapfig}
\usepackage{amssymb}
\usepackage{booktabs}
\usepackage{makecell}
\usepackage{subcaption}
\usepackage{xcolor}
\usepackage{colortbl}

\definecolor{mygray}{gray}{.9}
\newcommand\net{CL-Net}

\makeatletter
\DeclareRobustCommand\onedot{\futurelet\@let@token\@onedot}
\def\@onedot{\ifx\@let@token.\else.\null\fi\xspace}

\def\eg{\emph{e.g}\onedot}

\def\etal{\emph{et al}\onedot}
\makeatother

\makeatletter
\newcommand{\printfnsymbol}[1]{%
  \textsuperscript{\@fnsymbol{#1}}%
}
\makeatother

\begin{document}

\pagestyle{headings}
\mainmatter
\def\ECCVSubNumber{263}  

\title{Check and Link: Pairwise Lesion Correspondence Guides Mammogram Mass Detection} 


\titlerunning{CL-Net}
%
\author{Ziwei Zhao\thanks{Equal contribution.}\inst{1,4,5} \and
        Dong Wang$^\star$\inst{2} \and
        Yihong Chen\inst{1} \and \\
        Ziteng Wang\inst{3} \and
        Liwei Wang\inst{1,2}}
\authorrunning{Z. Zhao et al.}
%
\institute{Center for Data Science, Peking University\\
\email{zhaozw@stu.pku.edu.cn},
\email{chenyihong@pku.edu.cn} \\ 
\and
Key Laboratory of Machine Perception, MOE, School of Artificial Intelligence, Peking University \\
\email{wangdongcis@pku.edu.cn}, \email{wanglw@cis.pku.edu.cn} \\ \and 
Yizhun Medical AI Co., Ltd \\
\email{ziteng.wang@yizhun-ai.com} \\ \and
Pazhou Laboratory (Huangpu) \\ \and 
Peng Cheng Laboratory
}
\maketitle

\begin{abstract}

Detecting mass in mammogram is significant due to the high occurrence and mortality of breast cancer. In mammogram mass detection, modeling pairwise lesion correspondence explicitly is particularly important. However, most of the existing methods build relatively coarse correspondence and have not utilized correspondence supervision. In this paper, we propose a new transformer-based framework {\net} to learn lesion detection and pairwise correspondence in an end-to-end manner. In {\net}, View-Interactive Lesion Detector is proposed to achieve dynamic interaction across candidates of cross views, while Lesion Linker employs the correspondence supervision to guide the interaction process more accurately. 
The combination of these two designs accomplishes precise understanding of pairwise lesion correspondence for mammograms.
Experiments show that {\net} yields state-of-the-art performance on the public DDSM dataset and our in-house dataset. Moreover, it outperforms previous methods by a large margin in low FPI regime.

\end{abstract}

\section{Introduction}

With the highest incidence of cancers in women, breast cancer has become a serious threat to human health worldwide. In recent years, mammography screening has been used by most hospitals as a common examination for its effectiveness and non-invasiveness. Detecting mass is one of the core objectives for mammography screening since mass behaved spiculated and irregular is a typical sign of breast cancer. However, gland overlap and occlusion are great obstacles for distinguishing mass from the gland, accordingly identifying suspicious lesions on mammogram is difficult for both radiologists and deep learning models.

\begin{figure}[t]
\centering
\includegraphics[width=0.75\linewidth]{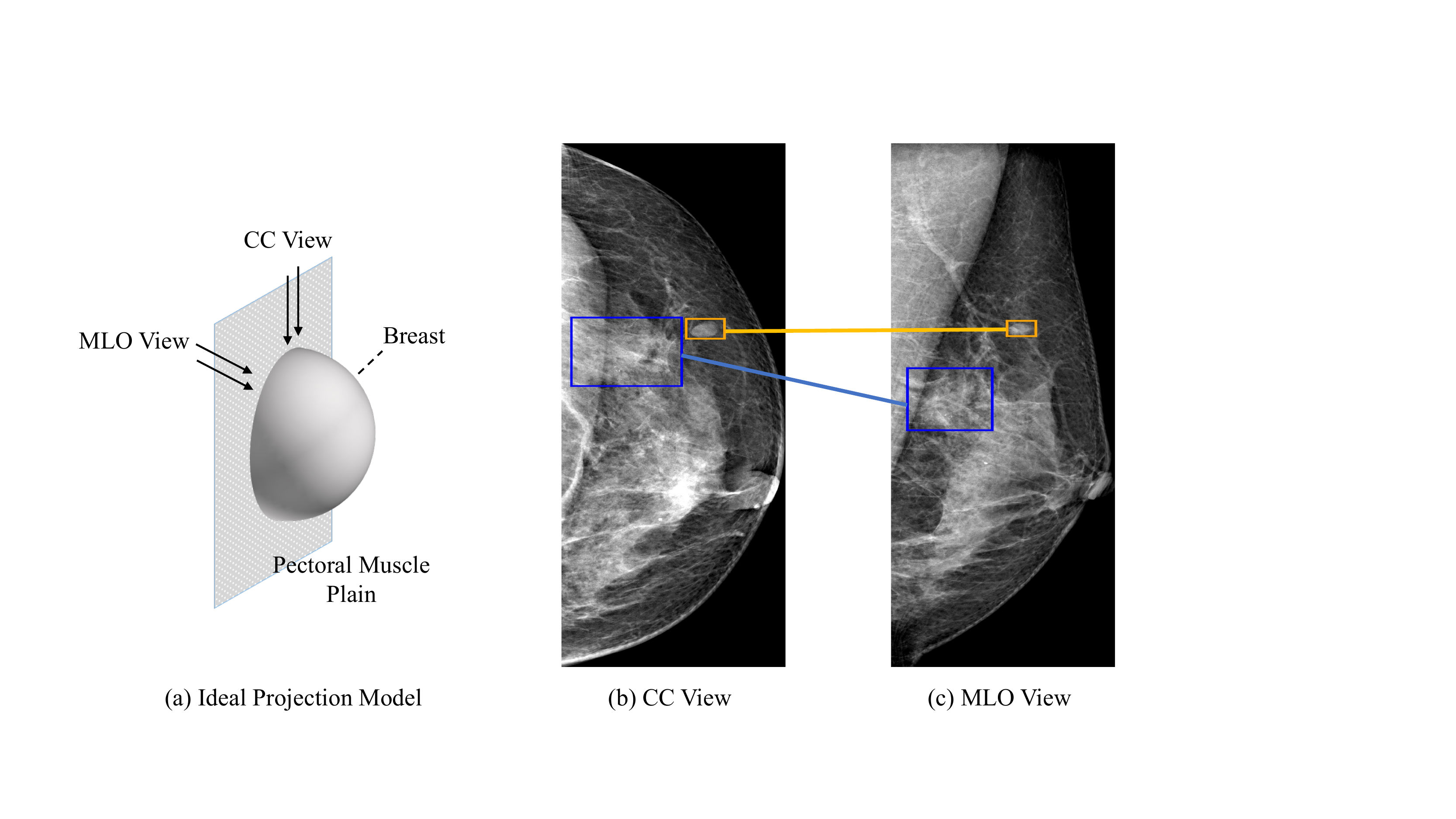}
\caption{(a) In mammograms, the CC view is a top-down view, while the MLO view is taken at a certain angle from the side. (b - c) show an example. Lesions connected by a line from different views are the projections of the same mass instance.}
\label{fig:mammo}
\end{figure}

In clinical practice, as shown in Figure~\ref{fig:mammo}, each breast is taken from two different angles, which are cranio-caudal (CC) view and mediolateral oblique (MLO) view, respectively. The complementary information of the ipsilateral view (CC view and MLO view of the same breast) will help radiologists to make better decisions for lesion detection. They usually cross-check the possible lesion locations in CC view and MLO view repeatedly. Once the relevant evidences are found in both views, the existence of the lesion can be confirmed. We call the co-existence of the same mass manifestations in both of the two views \textbf{pairwise lesion correspondence}. An example breast mammogram with two lesion pairs is shown in Figure~\ref{fig:mammo} (b-c). 

As for deep models, it is also particularly important to model the pairwise lesion correspondence explicitly for lesion detectors. 
Firstly, once the model is empowered with the ability to model pairwise lesion correspondences, the complementary information from the auxiliary view will help to distinguish the suspicious regions of the examined view, which is in line with the analysis logic of radiologists. 
Besides, the correspondences are also important supervision signals to train the network. The supervision of pairwise correspondences can guide the network to establish more accurate relations across the two views, which can further improve the detection performance.

\begin{figure}[t]
\centering
\includegraphics[width=0.75\linewidth]{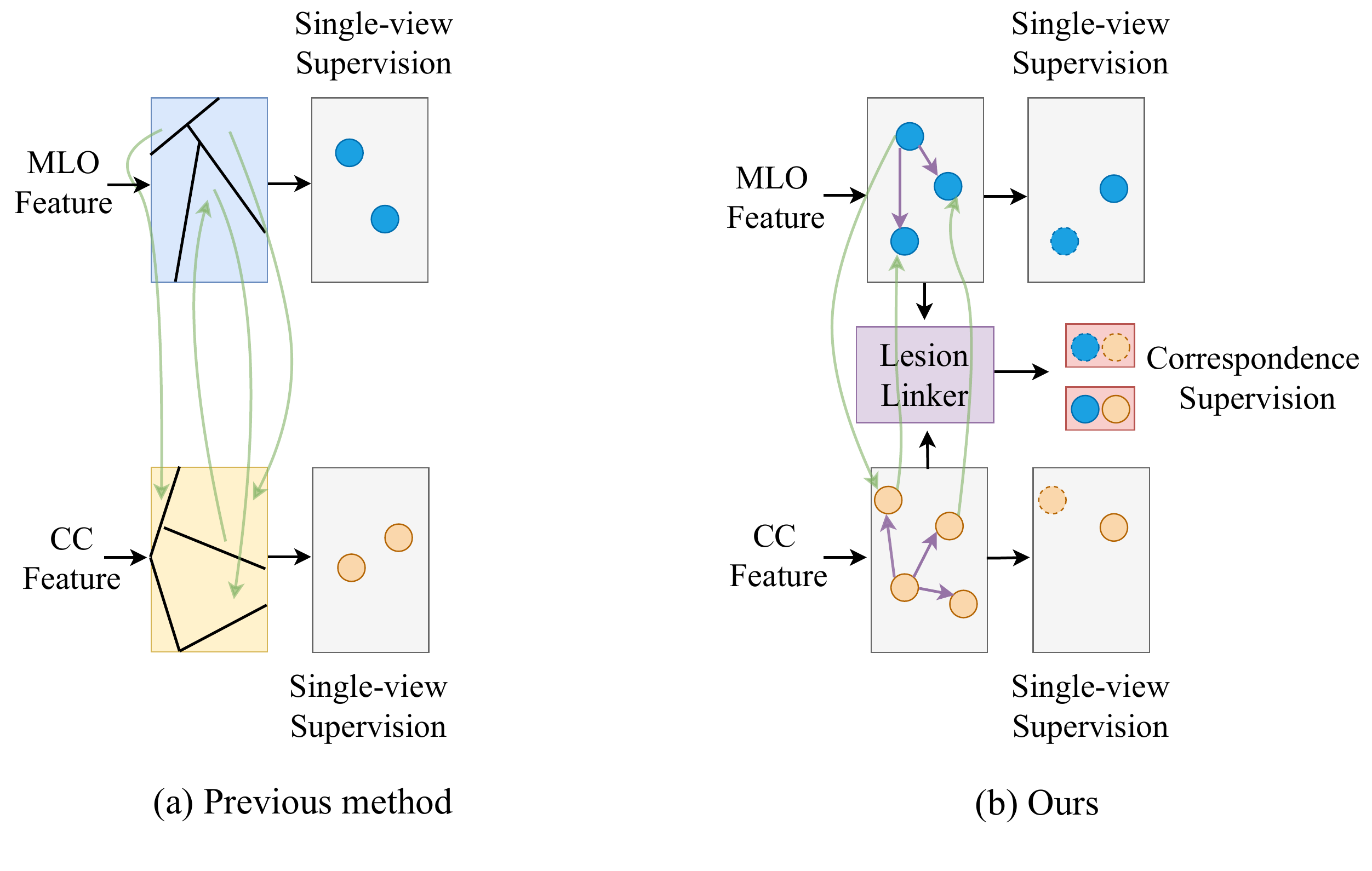}
\caption{(a) Previous methods~\cite{liu2020cross} split mammograms into several parts and model part-level relationships without correspondence supervision. (b) We establish lesion-level interaction across views, and leverage correspondence supervision to guide the network training procedure explicitly.}
\label{fig:model_sketch}
\end{figure}

Previous works have attempted to model lesion correspondences~\cite{ma2019cross,liu2020cross,liu2021compare,yang2021momminet}, however, the correspondence captured by these works is not accurate enough to represent pairwise lesion correspondence. For example, previous SOTA method BG-RCNN~\cite{liu2020cross} divides the image into multiple parts and builds part-wise correspondence using a graph neural network (Figure~\ref{fig:model_sketch}(a)), which leads to relatively coarse correspondence. Meanwhile, the utilization of correspondence supervision is not considered by previous works.

In this paper, we propose \textbf{\net} to model more precise pairwise lesion correspondence. We design a transformer-based network structure to learn the lesion detection and pairwise correspondence in an end-to-end manner.
As shown in Figure~\ref{fig:model_sketch}(b), our {\net} can not only build pairwise lesion correspondence for lesion candidates detected by single-view lesion detectors, but also leverage correspondence supervision to guide the network training procedure for discovering accurate and sensible pairwise lesion correspondence.

Specifically, we first propose \textbf{View-Interactive Lesion Detector} (VILD) to achieve dynamic interaction across lesion candidates of MLO and CC views. 
We build our model upon modern transformer-based object detectors (\eg DETR~\cite{carion2020end}).
These detectors often adopt the query mechanism, where each object query can be regarded as an abstract representation of a lesion candidate, and the information flow across queries is suitable for capturing the correspondences for lesion pairs. Therefore, we apply the inter-attention layer between the object queries' outputs of MLO and CC views to build relationships for the two views, which captures pairwise lesion correspondences in an elegant and efficient way.

Furthermore, we propose \textbf{Lesion Linker} to learn the precise pairwise lesion correspondence during network training. Lesion linker summarizes all lesion information from MLO and CC views by taking all lesion candidates generated by object queries as inputs, and then employs the \textbf{link query} and a decoder-like net structure to produce paired lesion outputs. Like DETR, we use a set prediction approach to output lesion pairs, where each pair is a point in the set. Hence, the lesion linker can also be trained with a set matching loss. 
Under the guidance of pair supervision, the inter-attention layer in VILD can obtain more direct and precise correspondence, which will benefit the training of the detector.

Experimentally, the results show that the proposed approach outperforms the previous SOTA methods by a large margin on both the public dataset DDSM~\cite{ddsm} and an in-house dataset. 
The ablation study validates the effectiveness of each part in our design.

In a nutshell, our contributions are three-folds:
\begin{itemize}
    \item To the best of our knowledge, our work is the first to model and learn pairwise lesion correspondence explicitly for mammogram mass detection, which is essential for cross-view reasoning.
    \item VILD and lesion linker are proposed to achieve precise lesion correspondence.
    \item We propose a novel framework, which achieves a new SOTA performance for mammogram mass detection with ipsilateral views and surpasses all previous methods by a large margin.
\end{itemize}

\section{Related Work}

\subsection{Mammogram Mass Detection}
Traditional approaches~\cite{campanini2004novel,eltonsy2007concentric,mudigonda2001detection,tai2013automatic} usually use complex preprocessing and design hand-crafted features for mammogram mass detection. However, due to the low representation ability, the performance of these methods is not satisfactory. In the past few years, deep learning has been introduced to this area. Most of works~\cite{cao2019deeplima,agarwal2019automatic,xi2018abnormality,ribli2018detecting} only use a single view for detection, while recently several studies~\cite{ma2019cross,liu2020cross,liu2021compare,yang2021momminet} attempt to establish cross-view reasoning mechanism for mammogram mass detection. Ma \etal~\cite{ma2019cross} and Yang \etal~\cite{yang2021momminet} use relation module~\cite{hu2018relation} to model the relationships of lesion proposals across views. Liu~\etal~\cite{liu2020cross} seeks to leverage bipartite graph convolutional network to achieve part-level correspondence. C2-Net~\cite{liu2021compare} preprocesses the mammograms for column-wise alignment and performs column-wise correspondence between cross-views, since they assume that the perpendicular distance to the chest of the same lesion in CC view and MLO view is roughly the same.
Although these methods model the correspondence of the two views to a certain extent, however, the correspondence is generated freely without any pairwise supervision. Perek~\etal~\cite{perek2018siamese} proposes a Siamese approach to achieve cross-view mass matching, while the performance of mass detection is not considered. Different from above approaches, our {\net} can model and learn the pairwise lesion correspondence explicitly, which significantly improves the detection performance.

\subsection{Object Detection and HOI Detection with Transformer}
Transformer~\cite{vaswani2017attention} has drawn great attention in computer vision recently~\cite{carion2020end,zhu2020deformable,dosovitskiy2020image,wang2021end}. In the area of object detection, the first representative of the transformer-based detector is DETR~\cite{carion2020end}. DETR employs a transformer encoder-decoder architecture with object queries to hit the instances in the images. It regards object detection as a set prediction task, and uses a set matching method~\cite{carion2020end} to train the network. Afterwards, Deformable DETR~\cite{zhu2020deformable} is proposed as a variant of DETR. Deformable DETR uses the local receptive fields for attention layers, which reduces computational complexity significantly and speeds up convergence. 
Moreover, DETR has also been appied to the task of Human-Object Interaction detection~\cite{kim2021hotr,chen2021reformulating,zou2021end,zhang2021mining}. Chen~\etal~\cite{chen2021reformulating} and Zou~\etal~\cite{zou2021end} reformulate HOI detection as a set prediction task and predict humans, objects and their interactions directly. HOTR~\cite{kim2021hotr} utilizes HO Pointers to associate the outputs of two parallel decoders, which leverages the self-attention mechanisms to exploit the contextual relationships between humans and objects. It is worth mentioning that HOI detection focuses on predicting the associations of humans and objects, while in this paper mammogram mass detection is evaluated by the detection results of each single image view. Different from HOI detection, the correspondence of MLO and CC views is regarded as the auxiliary supervision to promote the detection model. Our proposed lesion linker takes the advantage of this supervision to guide the training of VILD. 

\subsection{Learnable Image Matching}
The well-known image matching in computer vision aims to establish dense correspondences across images for camera pose recovery and scene structure estimation in geometric vision tasks, such as Structure-from-Motion (SfM) and Simultaneous Localization and Mapping (SLAM)~\cite{detone2018superpoint,dusmanu2019d2,ono2018lf,yi2018learning,ranftl2018deep,brachmann2019neural,sarlin2020superglue,sun2021loftr}. 
These methods rely on dense interest points as local descriptors to build pixel-to-pixel dense correspondences for multiple views.
However, in mammograms, the two views are two different projections of 3D breast, which means there is no precise pixel-to-pixel correspondence.
Therefore, we can only model the sparse lesion-to-lesion correspondence for accurate lesion detection. 
Compared with pixel-level matching, extracting the pairwise lesion correspondence is a high-level vision task that requires the network to understand lesion instances in advance. Therefore, we design the lesion linker and use link queries after the detector to learn lesion matching.

\begin{figure*}
\centering
\includegraphics[width=1.0\textwidth]{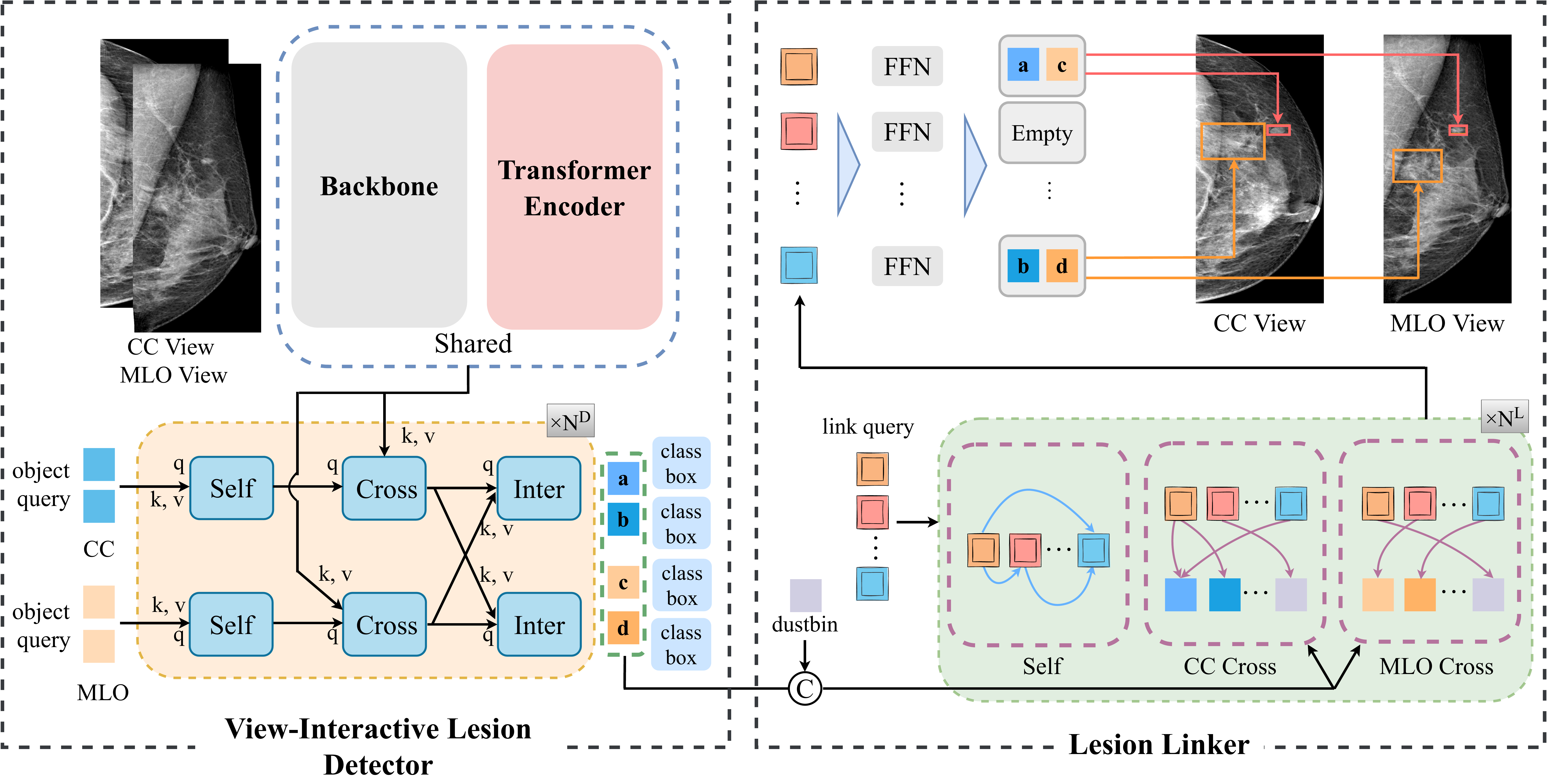}
\caption{Overview of our proposed {\net}. A pair of mammograms are firstly processed by View-Interactive Lesion Detector (VILD) to achieve dynamic interaction across lesion candidates of MLO and CC views. Then, the embeddings outputted by the last decoder layer in VILD are provided for Lesion Linker to learn the precise pairwise lesion correspondence by learnable link queries. }
\label{fig:framework}
\end{figure*}

\section{Methodology}

In this section, we will elaborate on the design of the proposed {\net}. The name CL stands that our method can Cross-\textbf{C}heck the two views and \textbf{L}ink the corresponding lesions across views.
An overview of the whole pipeline is illustrated in Figure~\ref{fig:framework}. To be specific, we first explain how the proposed View-Interactive Lesion Detector (VILD) along with the Lesion Linker establishes pairwise lesion correspondence. Then we discuss how to effectively train the network by presenting the training details including label assignment rules for lesion correspondence and the final loss function.

\subsection{Reviewing DETR}

\label{Sec. DETR}

Recently, DETR~\cite{carion2020end} has drawn great attention since it proposes a novel paradigm for object detection through transformer encoder-decoder architecture. It reformulates object detection as a set prediction task and adopts one-to-one label assignment between ground truth and predicted objects, which achieves an end-to-end object detector.

\textbf{Multi-head Attention.}
Attention is the core component in Transformer architecture. The standard version of attention can be written as follows:
\begin{eqnarray}
\text{Attention}(Q, K, V)=\text{softmax}(\frac{QK^T}{\sqrt{d_k}})V,
\end{eqnarray}
where $Q$, $K$, $V$ stand for query vector, key vector and value vector, respectively. $d_k$ is the vector dimension.

Multi-head attention is the extension of the standard version:
\begin{eqnarray}
&\text{MultiHeadAttn}(Q, K, V) = \text{Concat}(H_1, H_2, ..., H_m),\\
& H_i = \text{Attention}(QW_i^Q, KW_i^K, VW_i^V),    
\end{eqnarray}
where $m$ is the number of heads, $W^Q_i,W^K_i,W^V_i$ are projection matrices in the $i$-th head to map the original vector into a vector with lower dimension. For convenience, we use $\mathcal{M}$ to denote MultiHeadAttn in the following.

\textbf{Object Query.}
In DETR, object queries can be regarded as abstract representations of objects. After image features are extracted by the transformer encoder, queries will interact with these features through self-attention and cross-attention layers in the transformer decoder to aggregate instance information. Finally, several feed-forward network (FFN) layers are applied to decode box location and class information for each object query.

\subsection{View-Interactive Lesion Detector}

Dynamic interaction across lesion candidates of MLO and CC views is very helpful in establishing pairwise lesion correspondence. Therefore, we first propose the View-Interactive Lesion Detector (VILD) which aims to transfer lesion information across views effectively. The architecture of VILD is elaborated in the left part of Figure~\ref{fig:framework}. VILD is a transformer-based detector that also employs object query as an abstraction of object. VILD takes mammogram of MLO and CC views as input and passes them through the shared backbone and feature encoder to encode the image content. Afterwards, two sets of object queries (one for MLO and one for CC) are fed into a specially designed decoder to predict lesions' position and class for each view while taking the lesion information from the ipsilateral view into consideration.

To be specific, we append an additional inter-attention layer at the end of each transformer decoder block to achieve dynamic interaction across views. Object queries can be regarded as abstract representations of objects, thus directly applying cross-view inter-attention can be realized as an elegant and efficient way to capture pairwise lesion correspondence. 
The cross-view inter-attention is also instantiated as a multi-head attention block which takes intermediate embedding of one view as queries and intermediate embedding of the other view as keys and values. 
Formally, suppose the number of object queries of each view is $N$ and denote the embeddings output by cross-attention layer in the $i$-th decoder layer as $E^{c}_i,~E^{m}_i \in \mathbb{R}^{N \times D}$ for CC view and MLO view respectively, then the enhanced embeddings are obtained through attention mechanism which could be expressed as (take CC view for example),
\begin{align}
     & E^{c*}_i = E^{c}_i + \mathcal{M}(E^{c}_i + P^{c}, E^{m}_i + P^{m}, E^{m}_i),
\end{align}
where $P^{m}$ and $P^{c}$ denote the positional encodings for MLO and CC view's embedding, respectively. The positional encodings are learnable vectors, which are the same as Deformable DETR. $\mathcal{M}$ is MultiHeadAttn as defined in \ref{Sec. DETR}.
The enhanced embedding for MLO view is obtained vice versa:
\begin{align}
     & E^{m*}_i = E^{m}_i + \mathcal{M}(E^{m}_i + P^{m}, E^{c}_i + P^{c}, E^{c}_i).
\end{align}

By passing through the decoder layer for several times, cross-view lesion correspondence is gradually transferred and formed bidirectionally with the help of inter-attention block. This aligns with how radiologists identify lesions. They usually search for potential lesions in both views back and forth. Once a suspicious region is discovered in one view, they will check all possible positions in the other view in order to find the corresponding lesion with similar spatial and visual information. 

\begin{wrapfigure}[21]{r}{0.5\textwidth}
\centering
\includegraphics[width=1.0\linewidth]{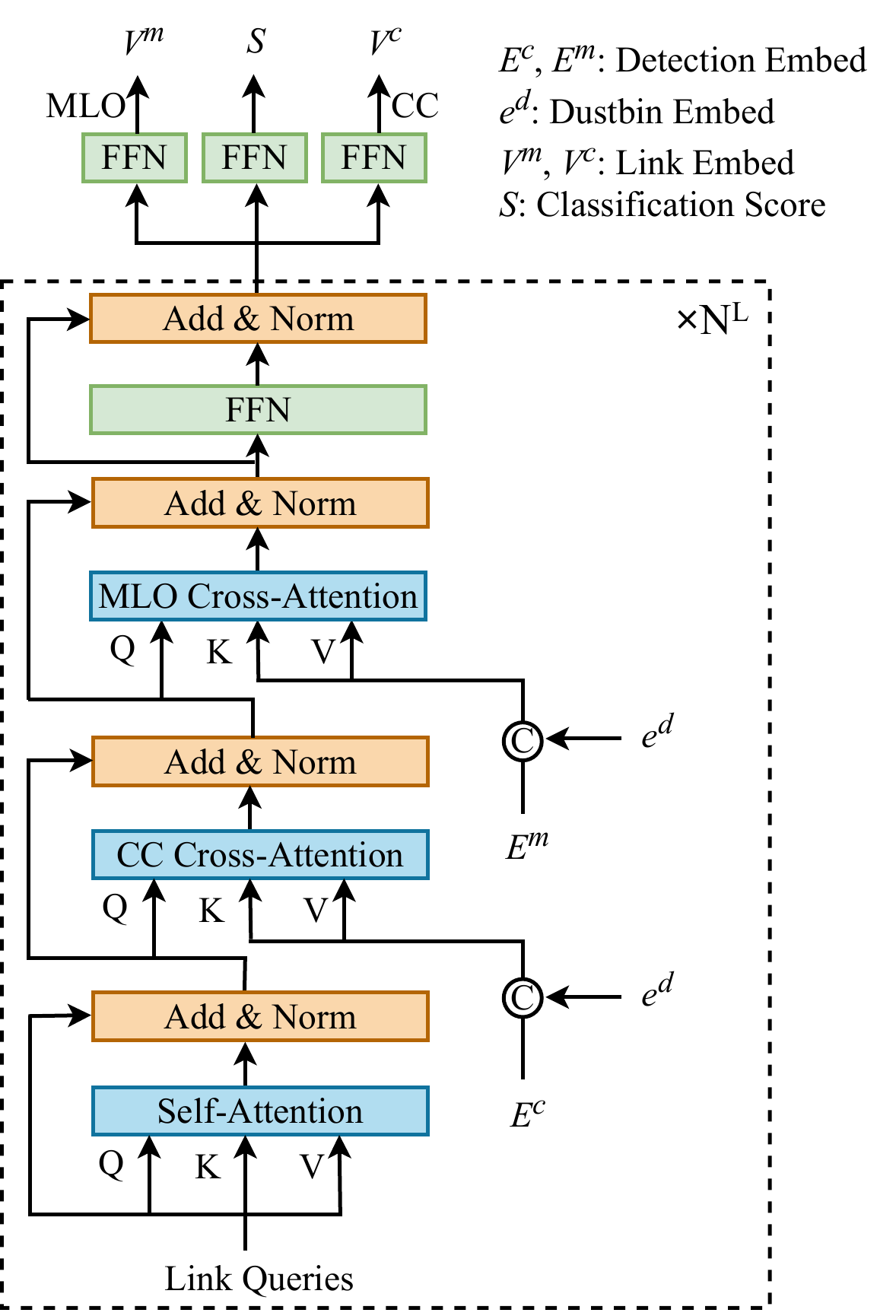}
\caption{Architecture of Lesion Linker.}
\label{fig: Lesion Linker}
\end{wrapfigure}

Denote the embeddings for CC view and MLO view outputted by the last decoder layer as $E^c,~E^m \in \mathbb{R}^{N \times D}$, the detection results are then predicted by FFN layers with $E^c$ and $E^m$ as input.

\subsection{Lesion Linker}
\label{sec: Lesion Linker}

In VILD, the establishment of cross-view dynamic interaction endows lesions from one view with the ability to form correspondence with lesions from the other view. We argue that by explicitly utilizing the guidance of the pair supervision, a more accurate pairwise lesion correspondence could be achieved and the detection ability of the network could be further boosted. We propose Lesion Linker, a transformer decoder-like structure to take full advantage of the pair supervision. The architecture of lesion linker is illustrated in the right part of Figure~\ref{fig:framework}. Lesion linker adopts link query, which is initialized as a set of learnable vectors, as abstract representations of possible pairwise relationships. Given output embeddings $E^c$ and $E^m$ from VILD as input, link queries will interact with them to extract lesion information and gradually focus on specific lesion pairs. Each link query will finally predict a triplet including link embeddings for CC and MLO views and lesion pair score through FFN layers. Given these embeddings, corresponding detection results in MLO view and CC view could be linked together to form pairwise lesion detection results. In the following, we will elaborate on the key designs of our lesion linker.

\textbf{Dustbin Embedding. } 
In clinical practice, mammogram is a projection along the X-ray direction in which lots of information is lost, thus some mass instances can only be seen in one view.
To cope with this special situation for lesion linker, we set a learnable vector $e^d \in \mathbb{R}^D$, named dustbin embedding. Detection embeddings that have no correspondence should be linked to it.

We concatenate detection embeddings from CC view and MLO view with dustbin embedding to obtain the complete version $\tilde{E}^{c}, \tilde{E}^{m} \in \mathbb{R}^{(N+1) \times D}$:
\begin{eqnarray}
    & \tilde{E}^{c}=\text{Concat}(E^c, e^d),\\  & \tilde{E}^{m}=\text{Concat}(E^m, e^d).
\end{eqnarray}

More explanations about dustbin embedding can be found in the supplementary materials.

\textbf{Architecture.} As illustrated in Figure~\ref{fig: Lesion Linker}, at first link queries $Q$ are passed through the multi-head self-attention layer. Then they will interact with detection embeddings from CC and MLO views sequentially to extract view-dependent information to form pairwise relationships. The process could be written as
\begin{align}
    \dot{Q} &= Q + \mathcal{M}(Q, Q, Q), \\
    \ddot{Q} &= \dot{Q} + \mathcal{M}(\dot{Q}, \tilde{E}^{c} + P^c, \tilde{E}^{c}), \\
    \hat{Q} &= \ddot{Q} + \mathcal{M}(\ddot{Q}, \tilde{E}^{m} +P^m, \tilde{E}^{m}),
\end{align}
where $P^c$ and $P^m$ denote the positional encoding for $\tilde{E}^{c}$ and $\tilde{E}^{m}$.  Finally, $\hat{Q}$ is processed by a FFN layer to further enhance the representative ability. Above layers can be stacked for several times. Link queries transformed by stacked attention have fully explored lesion-level relationships from MLO view and CC view and the pairwise lesion correspondence is gradually formed. 

Motivated by~\cite{kim2021hotr}, at the top of lesion linker, we decode the correspondence by applying three FFN layers to predict the link embedding for CC and MLO views $V^c \in \mathbb{R}^{M \times D}$, $V^m \in \mathbb{R}^{M \times D}$ and lesion pair classification score $S \in \mathbb{R}^{M \times 1}$, respectively. $M$ is the number of the link queries and $D$ is the feature dimension. 
The predicted link embeddings $V^c$ and $V^m$ are used for indexing the detection results, which will be introduced later.
The classification score $S$ denotes the confidence that whether the pair of detection results captured by the link query is true positive.

\textbf{Lesion Correspondence Extracting.}
The output of lesion linker can be reformulated as a set of $M$ triplets, $\{\langle v_i^c, v_i^m, s_i\rangle\}_{i=1}^M$, where $v_i^c, v_i^m \in \mathbb{R}^D $ and $s_i \in \mathbb{R}^1 $ are the i-th row of $V^c, V^m$ and $S$. The pairwise lesion correspondence could be explicitly established by first calculating the feature similarity between detection embeddings $\tilde{e}_j^t \in \mathbb{R}^D$ and link embeddings $v_i^t$ for each view and then taking the index of the detection embedding with the highest similarity as result. Here $t \in \{c, m\}$ denotes CC view or MLO view, and $\tilde{e}_j^t$ is the j-th row of $\tilde{E}^t$. Formally, this process could be expressed as
\begin{equation}
    c_i =\mathop{\arg\max}\limits_{j}( \text{sim}(v_i^c, \tilde{e}_j^c)), \ 
    m_i =\mathop{\arg\max}\limits_{j}( \text{sim}(v_i^m, \tilde{e}_j^m)),
\end{equation}
where we use cosine similarity to measure the feature similarity:
\begin{align}
    \label{sim}
    \text{sim}(x, y) = \frac{x^T y}{||x||_2||y||_2}.
\end{align}

Finally, for each link query $q_i$, we could obtain its extracted lesion correspondence pair $\langle c_i,~m_i\rangle$ as result. Next we will discuss how to effectively train our network.

\subsection{Training Details}
We will elaborate on training details of our proposed {\net} in this section. To be specific, we first explain the label assignment rule for pairwise lesion correspondence. Then we introduce the loss function of our {\net}.

\textbf{Label Assignment for Lesion Correspondence.}
In original DETR, a one-to-one label assignment based on bipartite matching is used to assign training targets for the predicted bounding boxes. In our {\net}, we also aim to establish a similar rule to assign the pairwise ground truth lesion boxes to the set of link triplets predicted by lesion linker.

Our VILD shares a similar structure and training strategy with DETR, therefore through the label assignment rule for detection, we can obtain the assignment relationships between ground truth boxes and detection embeddings. 
Thus the pairwise ground truth boxes can be naturally converted to pairwise detection embeddings. We denote the conversion results as $y = \langle e^c, e^m, a=1 \rangle$. $e^c$ and $e^m$ denote the detection embedding converted from ground truth boxes for CC view and MLO view. 
For a lesion that can only be viewed in CC view, the converted result is $y = \langle e^c, e^d, a=1 \rangle$, in which $e^d$ denotes the dustbin embedding defined in Section \ref{sec: Lesion Linker}. The same is for the lesion that can only be viewed in MLO view.

Suppose the number of unique ground truth lesions is $K$. Then the set of converted lesion triplets from the ground truth lesions could be denoted as $Y=\{y_i\}_{i=1}^K$. The set of $M$ predictions from lesion linker could be similarly denoted as $\hat{Y}=\{\hat{y}_j = \langle v_j^c, v_j^m, s_j \rangle\}_{j=1}^M$. Since $K$ is less than $M$ in mammogram, we pad the ground truth set $Y$ with $\langle \varnothing, \varnothing, a=0 \rangle $ (no lesion pair) to the size of $M$, similar to DETR. We aim to find an optimal bipartite matching between these two sets by searching for a permutation of $M$
elements $\pi \in \Pi_M $ with the lowest cost:
\begin{align}
    \hat{\pi}=\mathop{\arg\min}\limits_{\pi \in \Pi} \sum_{i=1}^M \mathcal L_{\text{match}}(y_i, \hat{y}_{\pi(i)}),
\end{align}
where $\mathcal L_{\text{match}}$ is a matching cost between ground truth $y_i$ and prediction $\hat{y}_{\pi(i)}$. We consider two aspects when calculating the matching cost, which are the prediction scores and the similarity of ground truth embeddings and predicted link embeddings:
\begin{equation}
\begin{aligned}
\label{cost function}
     \mathcal{L}_{\text{match}}(y_i, \hat{y}_{\pi(i)}) 
     =  - \mathbf{1}_{\{a_i \neq 0 \}} \cdot [\mathcal{L}_{\text{emd}}(i,\pi(i))]^\alpha \cdot [\mathcal{L}_{\text{score}}(i,\pi(i))]^{1-\alpha},
\end{aligned}
\end{equation}
where $\mathcal{L}_{\text{emd}}$ and $\mathcal{L}_{\text{score}}$ denote cost of feature similarity and classification score. The operation of + 1 in Eq.~\ref{embed loss} aims to guarantee that $\mathcal{L}_{\text{emd}}$ is positive.
\begin{align}
\label{embed loss}
    &\mathcal{L}_{\text{emd}}(i,j) =\beta\text{sim}(e_i^c, v_j^c) + (1-\beta)\text{sim}(e_i^m, v_j^m) + 1, \\
    &\mathcal{L}_{\text{score}}(i,j) = s_{j}.
\end{align}

We adopt the weighted geometric mean of the feature similarity $\mathcal{L}_{\text{emd}}$ and classification score $\mathcal{L}_{\text{score}}$, in which $\alpha \in [0,1]$ is the balance hyper-parameter. The ablation study of the cost function and analysis can be found in Table~\ref{tab: cost function}. $\beta$ is set to 0.5 by default to adjust the ratio of feature similarity in CC view and MLO view. The optimal bipartite assignment can be obtained through the Hungarian algorithm efficiently as in~\cite{stewart2016end}.

\textbf{Training Loss.}
The final loss function can be written as follows:
\begin{align}
    \mathcal{L} = \mathcal{L}_\text{D} + \mathcal{L}_\text{Link},
\end{align}
where $\mathcal{L}_\text{D}$ is the loss function in DETR, $\mathcal{L}_\text{Link}$ is defined as
\begin{align}
    \mathcal{L}_\text{Link} = \sum_{i=1}^M[\mathbf{1}_{\{a_i \neq 0 \}}\lambda_{\text{sim}} \mathcal{L}_{\text{sim}}(i, \pi(i)) +  \lambda_{\text{cls}} \mathcal{L}_{\text{cls}}(a_i, s_{\pi(i)} )],
\end{align}
where $\lambda_{\text{sim}}$ and $\lambda_{\text{cls}}$ are weight hyper-parameters.
We adopt focal loss~\cite{lin2017focal} as the loss function $\mathcal{L}_{\text{cls}}$ for lesion pair classification.

Following~\cite{kim2021hotr}, we first calculate the similarity scores $S^t \in \mathbb{R}^{N+1}$, where $t \in \{c, m\}$ denotes the CC view and MLO view, and the j-th item of $S^t$ is $\text{sim}(v_{\pi(i)}^t, \tilde{e}_j^t)$. Then, we use Cross-Entropy Loss to localize the ground truth embeddings:
\begin{equation}
\begin{aligned}
    \mathcal{L}_{\text{sim}}(i, \pi(i))
    =\text{CrossEntropyLoss}(S^c, i) + \text{CrossEntropyLoss}(S^m, i).
\end{aligned}
\end{equation}

\subsection{Discussion of Match Learning Strategy}
\label{sec:discuss}

Our lesion linker learns the paired relationships of lesions by learning to predict MLO and CC embeddings which are close to the corresponding lesion pairs.
Our match learning strategy is a soft way, which gradually pushes the link embeddings to get closer to the ground-truth embeddings during training. 
Although there are also other alternative approaches for this task, learning lesion matching is not trivial. 
In this subsection, we compare our method with two other seemingly reasonable solutions, to strengthen the advantages of our method.

\textbf{Pair Verification. }
A straightforward solution to predict match pairs is to verify whether every two lesions from ipsilateral views are truly paired lesions. Following this design, we need to output a 2D matrix that represents the match probabilities of all possible pairs. The shape of the matrix should be $N \times N$, where $N$ is number of lesion candidates per view. However, since the number of possible lesion pairs is much larger than the number of truly paired lesions, it is difficult to extract useful training signals from such a small amount of pairwise annotation information.

Compared to the verification approach, the introduction of link queries decouples the pairwise training from the number of object queries $N$. The number of link queries $M$ could be in the same order of magnitude as $N$, thus the pair supervision signal could be fully utilized, leading to an easier optimization process.

\textbf{Paired Lesion Query. }
Another seemingly straightforward way is to predict pairwise lesions with query mechanism directly. With this paired lesion query, the network can output a pair of detected boxes in the two views for each query. Then, the form of outputted lesion pairs is similar to the extracted lesion pairs of lesion linker. Therefore, we can also adopt a similar set matching loss to train the network. With the paired lesion query, the object query for each view is not required anymore.

However, the optimization of paired lesion query is much harder than our lesion linker. Our {\net} first detects lesion candidates from each view (in VILD), therefore the lesion linker only focus on extracting the pairwise correspondence. While using the paired lesion query, the detection of objects and pairing are performed in the same step, which increases the difficulty of network training and results in inferior performance.

We elaborate on the implementation details of above two methods in the supplementary materials.
The experimental results are presented in section \ref{sec:ablation}.

\section{Experiments}

\subsection{Implementation Details}

Our model is based on Deformable DETR~\cite{zhu2020deformable} for its flexibility and fast convergence. We adopt ResNet-50~\cite{he2016deep} pre-trained from ImageNet~\cite{deng2009imagenet} as backbone. The number of object queries $N$ and link queries $M$ are set to 125 and 16, respectively. The loss weights $\lambda_{\text{sim}}$ and $\lambda_{\text{cls}}$ are 0.125 and 1.0 by default. We set $\alpha=0.5$ and $\gamma=2.0$ for the focal loss $\mathcal{L}_{\text{cls}}$. It is worth mentioning that since we mainly focus on the task of lesion detection, the final predictions come from VILD in the inference process.

We implement our network with PyTorch~\cite{adam19torch}. 
We train the network in an end-to-end manner on 8 GPUs for 25k iterations. For each GPU, we use 4 images containing two mammogram pairs.
Following Deformable DETR~\cite{zhu2020deformable}, we train our model using AdamW Optimizer~\cite{loshchilov2017decoupled} with base learning rate of $2\times10^{-4}$. We use the same multiplied factors for learning rates as~\cite{zhu2020deformable}, while the learning rates of lesion linker parameters are multiplied by 0.25. In addition, we adopt cosine learning rate schedule with warm-up. To avoid overfitting, we use several data augmentation methods (random flip, random crop, random normalization) in training.

\begin{table}[t]
\caption{Comparison with baselines and previous SOTA on DDSM dataset (\%).} 
\centering
\label{tab:sota1}
\scalebox{0.82}[0.82]{
\begin{tabular}{cccccc}
\Xhline{1.0pt}
\rowcolor{mygray}
\textbf{Method} & R@0.25 & R@0.5 & R@1.0 & R@2.0 & R@4.0  \\ \hline \hline
Mask RCNN~\cite{liu2020cross} & - & 76.0 & 82.5 & 88.7 & 91.4\\
Mask RCNN, DCN~\cite{liu2020cross} & - & 76.7 & 83.9 & 89.4 & 91.8\\
Deformable DETR~\cite{zhu2020deformable}  &73.8 &78.4 &83.7 &88.7 &93.7 \\
BG-RCNN~\cite{liu2020cross} & - & 79.5 & 86.6 & 91.8 & 94.5\\
\hline
\textbf{{\net}} &\textbf{78.1} & \textbf{83.1} & \textbf{88.0} & \textbf{92.4} & \textbf{95.0} \\
\Xhline{1.0pt}
\end{tabular}
}
\end{table}

\begin{table}[t]
\caption{Comparison with previous works on DDSM dataset (\%).} \label{tab:sota2}
\centering
\scalebox{0.82}[0.82]{
\begin{tabular}{cccc}
\Xhline{1.0pt}
\rowcolor{mygray}
\textbf{Method} & & R@t & \\ \hline \hline
Campanini \etal~\cite{campanini2004novel} & & 80@1.1 &\\
Eltonsy \etal~\cite{eltonsy2007concentric} & & 92@5.4, 88@2.4, 81@0.6 &\\
Sampat \etal~\cite{sampat2008model} & & 88@2.7, 85@1.5, 80@1.0 &\\
CVR-RCNN~\cite{ma2019cross} & & 92@4.4, 88@1.9, 85@1.2 &\\
\hline
\textbf{{\net}} & & \textbf{96@4.4, 92@1.9, 89@1.2}\\
\Xhline{1.0pt}
\end{tabular}
}
\end{table}

\subsection{Datasets}
We conduct experiments on the public DDSM dataset and our in-house dataset.

\textbf{DDSM dataset.}
DDSM~\cite{ddsm} is a widely used public dataset. It contains 2620 patient cases, and each case has four images, including MLO view and CC view for both breasts. We use the same data split method with previous studies~\cite{liu2020cross,ma2019cross,sampat2008model,campanini2004novel}. 
The original dataset does not provide lesion correspondence annotations, hence we fulfill the annotations with experienced radiologists.

\textbf{In-house dataset.}
We collect an in-house mammography dataset with 3,160 cases. Each case is annotated by at least two experts.  We randomly split the dataset into train, validation, and test set with the ratio of 8:1:1.

\textbf{Evaluation Metric.} We report recall ($R$) at $t$ false positives per image (FPI) to evaluate the performance following~\cite{liu2020cross,ma2019cross}. The metric can be simplified as $R@t$.

\begin{table}[t]
\caption{Ablation study on different components of {\net} on DDSM dataset (\%). VILD: View-Interactive Lesion Detector. LL: Lesion Linker. “not using VILD” means we use Deformable DETR directly.}
\centering
\label{tab: decoders}
\scalebox{0.82}[0.82]{
\begin{tabular}{ccccccc}
\Xhline{1.0pt}
\rowcolor{mygray}
  \textbf{VILD} & \textbf{LL} & R@0.25 & R@0.5 & R@1.0 & R@2.0  & R@4.0  \\ \hline \hline
& &73.8 &78.4 &83.7 &88.7 &93.7  \\
 \checkmark  & & 76.1  & 81.7  & 86.4  & 91.7  & 94.4 \\
 & \checkmark &  74.1 & 80.1  & 86.0  & 88.7 & 93.4 \\
\checkmark & \checkmark &\textbf{78.1} & \textbf{83.1} & \textbf{88.0} & \textbf{92.4} & \textbf{95.0} \\
\Xhline{1.0pt}
\end{tabular}
}
\end{table}

\subsection{Compare with State-of-the-art Methods}
We compare our methods with previous works on DDSM dataset in Table~\ref{tab:sota1} and Table~\ref{tab:sota2}. 
In Table~\ref{tab:sota1}, the results of Mask RCNN, Mask RCNN DCN, and BG-RCNN are from~\cite{liu2020cross}, and Deformable DETR is implemented by ourselves. 
In Table~\ref{tab:sota2}, we use the same FPIs as in CVR-RCNN~\cite{ma2019cross} and compare our method with previous works. 
From these two tables, we can draw a conclusion that our \net{} outperforms all baselines by a large margin and surpasses previous SOTA~\cite{liu2020cross,ma2019cross}. The performance of our method is more significant in low FPIs, outperforming BG-RCNN~\cite{liu2020cross} by 3.6 at $R@0.5$, which could benefit clinical practice a lot. The results on the in-house dataset are reported in the supplementary materials. Similar improvement over baselines on in-house dataset also demonstrates the superiority of our approach.

\subsection{Ablation Study}
\label{sec:ablation}

In this section, we elaborate on ablation studies for {\net}. Other ablation experiments are presented in the supplementary materials.

\textbf{Different Components of {\net}.}
We ablate the impact of different components of {\net} on detection performance in Table~\ref{tab: decoders}. There are mainly two important modules in {\net}, VILD and lesion linker. As shown in the table, using VILD can significantly improve the detection performance, while the improvement of employing lesion linker alone is marginal. 
Considering that learning accurate correspondences relies on the expression ablitily of VILD, it is explainable that the contribution of lesion linker is limited without VILD. The effect of lesion linker in {\net} is guiding the interaction process more precisely in the inter-attention layer of VILD. The experimental results also verifies our conjecture. The joint contributions of VILD and lesion linker improve the detection performance of VILD significantly ($+2.0$ at $R@0.25$).

\textbf{Different Strategies for Match Learning.}
We present the results of different strategies for match learning in Table~\ref{tab: different methods}. The methods described in section~\ref{sec:discuss} are adopted.
Pair verification method yields similar performance as VILD solely, which indicates that it is hard for the model to mine useful correspondences from plenty of feasibilities.
In addition, paired lesion query performs much worse than VILD ($-7.3$ at $R@0.25$) due to the difficulty of optimization. 
Our {\net} achieves the best performance attributing the success to the design of link query.

\begin{table}[t]
\begin{minipage}{0.45\textwidth}
\centering
\caption{Different strategies for match learning on DDSM dataset (\%). PV: Pair Verification. PL Query: Paired Lesion Query.} 
\label{tab: different methods}
\scalebox{0.82}[0.82]{\begin{tabular}{ccccc}
\Xhline{1.0pt}
\rowcolor{mygray}
  \textbf{Method} & R@0.25 & R@0.5 & R@1.0 & R@2.0   \\ \hline \hline
VILD & 76.1  & 81.7  & 86.4  & 91.7  \\ \hline \hline
PV &  75.7 & 82.1  & 86.4  & 90.7  \\
PL Query & 68.8  & 75.1  & 81.7  & 87.0  \\
\textbf{{\net}} &\textbf{78.1} & \textbf{83.1} & \textbf{88.0} & \textbf{92.4} \\
\Xhline{1.0pt}
\end{tabular}
}
\end{minipage}
\hfil
\hfil
\begin{minipage}{0.45\textwidth}
\centering
\caption{Ablation study on cost function of label assignment. Default parameter is marked by *.} 
\label{tab: cost function}
\scalebox{0.82}[0.82]{
\begin{tabular}{c | c | c c c c c}
\Xhline{1.0pt}
\rowcolor{mygray}
  \textbf{Method} & $\alpha$ & R@0.25 & R@0.5 & R@1.0 & R@2.0  \\ \hline \hline
 & 0.25 & 77.7 & 82.7 & 87.7 & 90.4 \\
  Add & 0.5 & 73.8 & 81.1 & 86.4 & 90.0 \\
   & 0.75 & 75.1  & 80.1 & 86.7 & 91.7  \\
  \hline
   & 0.25 & 76.1 & \textbf{83.7} & \textbf{89.7} & 91.7  \\
 Mul & 0.5* &  \textbf{78.1} & 83.1 & 88.0 & \textbf{92.4} \\
  & 0.75 & 74.8 & 80.4 & 85.4 & 90.0 \\
\Xhline{1.0pt}
\end{tabular}
}
\end{minipage}
\end{table}

\textbf{Cost Function of Label Assignment.}
We investigate the effect of different forms of cost function for label assignment on our model in Table~\ref{tab: cost function}. Method `Mul' denotes the cost function in Eq.~\ref{cost function}, while `Add' refers to the weighted sum of $\mathcal{L}_{\text{emd}}$ and $\mathcal{L}_{\text{score}}$, where $\alpha$ is also the weighting factor.
The experimental results show that method `Mul' achieves better performance than `Add', which could be mainly attributed to the sensitivity to both $\mathcal{L}_{\text{emd}}$ and $\mathcal{L}_{\text{score}}$ in the form of multiplication. 

The visualization and error analysis can be found in the supplementary materials.

\section{Conclusion}
In this work, we present {\net}, a novel mammogram mass detector based on transformer architecture. 
Our {\net} can not only model precise pairwise lesion correspondence, but also leverage correspondence supervision to guide the network training.
The experimental results conducted on the public DDSM dataset and an in-house dataset show that {\net} surpasses the state-of-the-art methods by a large margin.

\noindent \textbf{Acknowledgement} This work is supported by Exploratory Research Project of Zhejiang Lab (No. 2022RC0AN02), Project 2020BD006 supported by PKUBaidu Fund.

\appendix
\newpage
\section{Appendix}

\subsection{Performance on the In-house Dataset}

\begin{table}
\caption{Performance on the in-house dataset (\%).}
\centering
\label{tab:sota3}
\scalebox{0.82}[0.82]{
\begin{tabular}{cccccc}
\Xhline{1.0pt}
\rowcolor{mygray}
\textbf{Method} &R@0.125& R@0.25 & R@0.5 & R@1.0 & R@2.0\\ \hline \hline
Faster RCNN~\cite{ren2015faster} & 78.8 & 84.0 & 88.6 & 91.2 & 93.6\\
Mask RCNN~\cite{he2017mask} & 80.5 & 85.2 & 89.2 & 92.1 & 94.8\\
Deformable DETR~\cite{zhu2020deformable} & 79.2 & 84.7 & 88.6 & \textbf{93.4} & 95.3 \\
\hline
\textbf{{\net}} & \textbf{82.2} & \textbf{87.8} & \textbf{90.6} & \textbf{93.2} & \textbf{96.0} \\
\Xhline{1.0pt}
\end{tabular}
}
\end{table}

\subsection{Additional Ablation Study}

\begin{table} 
\begin{center}
\vspace{-30pt}
\caption{Ablation study on number of attention layers in lesion linker.} 
\label{tab: num of layers}
\scalebox{1.0}[1.0]{
\begin{tabular}{c c c c c c c}
\Xhline{1.0pt}
\rowcolor{mygray}
  \textbf{Number} & R@0.25 & R@0.5 & R@1.0 & R@2.0  & R@4.0  \\ \hline \hline
 1 & 77.1 & 81.4 & \textbf{88.4} & \textbf{92.4} & 94.4 \\
  2 & 77.1  & 82.4   & 88.0   &  91.4  & 93.4 \\
 3 &  \textbf{78.1} & \textbf{83.1} & 88.0 & \textbf{92.4} & \textbf{95.0}   \\
 4 & 75.1 & 80.4 & 86.4 & 90.7 & 93.0 \\
 5 & 75.1 & 80.4 & 85.0 & 90.0 & 92.0 \\
\Xhline{1.0pt}
\end{tabular}
}
\vspace{-20pt}
\end{center}
\end{table}

\textbf{Number of Attention Layers in Lesion Linker.}
Since several attention layers can be stacked in Lesion Linker for more powerful reasoning ability, we explore the effect of the number of attention layers in Table~\ref{tab: num of layers}. From the table, with the number of attention layers increasing from 1 to 3, performance of {\net} continues to improve and gets the best overall recalls at 3. Thus the number of layers is set to 3 as our default setting.

\subsection{Details of Match Learning Strategies}
In the main text of our paper, in order to achieve precise pairwise lesion correspondence, lesion linker utilizes correspondence supervision to guide the interaction process across lesion candidates of MLO and CC views. Besides lesion linker, we also introduce two alternative approaches for pairwise correspondence learning in Section 3.5 of the main paper to highlight the advantages of our proposed method. We will elaborate on the details of the alternative approaches in the following.

\subsubsection{Pair Verification}
Verifying whether each two lesion candidates from ipsilateral views are truly paired lesions is a straightforward method to achieve lesion matching. We instantiate this model based on View-Interactive Lesion Detector (VILD). Given output embeddings $E^c \in \mathbb{R}^{N \times D}$ and $E^m \in \mathbb{R}^{N \times D}$ from VILD as inputs, a MLP layer firstly transforms them to new representation space which can be expressed as:
\begin{eqnarray}
    & {E}^{c*}=\text{MLP}(E^c),\ {E}^{m*}=\text{MLP}(E^m).
\end{eqnarray}
Similar with lesion linker, we also set a learnable dustbin embedding $e^d \in \mathbb{R}^D$ to obtain the complete version $\hat{E}^c$, $\hat{E}^m \in \mathbb{R}^{(N+1) \times D}$:
\begin{eqnarray}
    & \hat{E}^{c}=\text{Concat}(E^{c*}, e^d),\ \hat{E}^{m}=\text{Concat}(E^{m*}, e^d).
\end{eqnarray}
We construct the 2D matching matrix which represents the match probabilities of all possible pairs by calculating the similarity score for every two embeddings from ipsilateral views:
\begin{eqnarray}
    S_{i,j} = \langle\hat{E}^{c}_i, \hat{E}^{m}_j\rangle,
\end{eqnarray}
where $\langle\cdot,\cdot\rangle$ is the inner product. $(i, j) \in [1, N+1] \times [1, N+1]$. $\hat{E}^{t}_i$ denotes the i-th row of $\hat{E}^{t}$, and $t \in \{c, m\}$. 

To learn pairwise lesion correspondence in a supervised manner, we introduce the ground truth matching matrix $M \in \mathbb{R}^{(N+1) \times (N+1)}$. Through the label assignment rule for detection in DETR~\cite{carion2020end}, pairwise ground truth boxes can be naturally converted to pairwise ground truth embeddings. We denote the ground truth match ids for the embeddings of ipsilateral views as $\mathcal{GT}=\{(\mu_i, v_i)\}_{i=1}^n$, which represent that the $\mu_i$-th embedding in CC view and the $v_i$-th embedding in MLO view are a pair of ground truth embeddings. If a lesion corresponding to the $\mu_i$-th embedding can only be viewed in CC view, we denote its match id as $(\mu_i, N+1)$. 
Lesions that can only be viewed in MLO view can also be processed in a similar way.
Thus $M$ can be obtained as follows:

\begin{equation}
M[i, j] = \left\{
\begin{aligned}
& 1, \ \ \ \text{if}\   (i, j) \in \mathcal{GT} \\
& 0, \ \ \ \text{else}
\end{aligned}
\right.
\end{equation}
 
The final loss function for this method can be written as follows:
\begin{eqnarray}
    \mathcal{L} = \mathcal{L}_\text{D} + \mathcal{L}_{\text{V}}(S, M),
\end{eqnarray}
where $\mathcal{L}_D$ is the loss function in DETR, and $\mathcal{L}_{\text{V}}$ is used to supervise the matching results. We adopt focal loss~\cite{lin2017focal} as the loss function $\mathcal{L}_{\text{V}}$ to achieve the supervision between predicted matching matrix $S$ and ground truth matching matrix $M$.

\begin{figure}[t]
\centering
\includegraphics[width=0.45\linewidth]{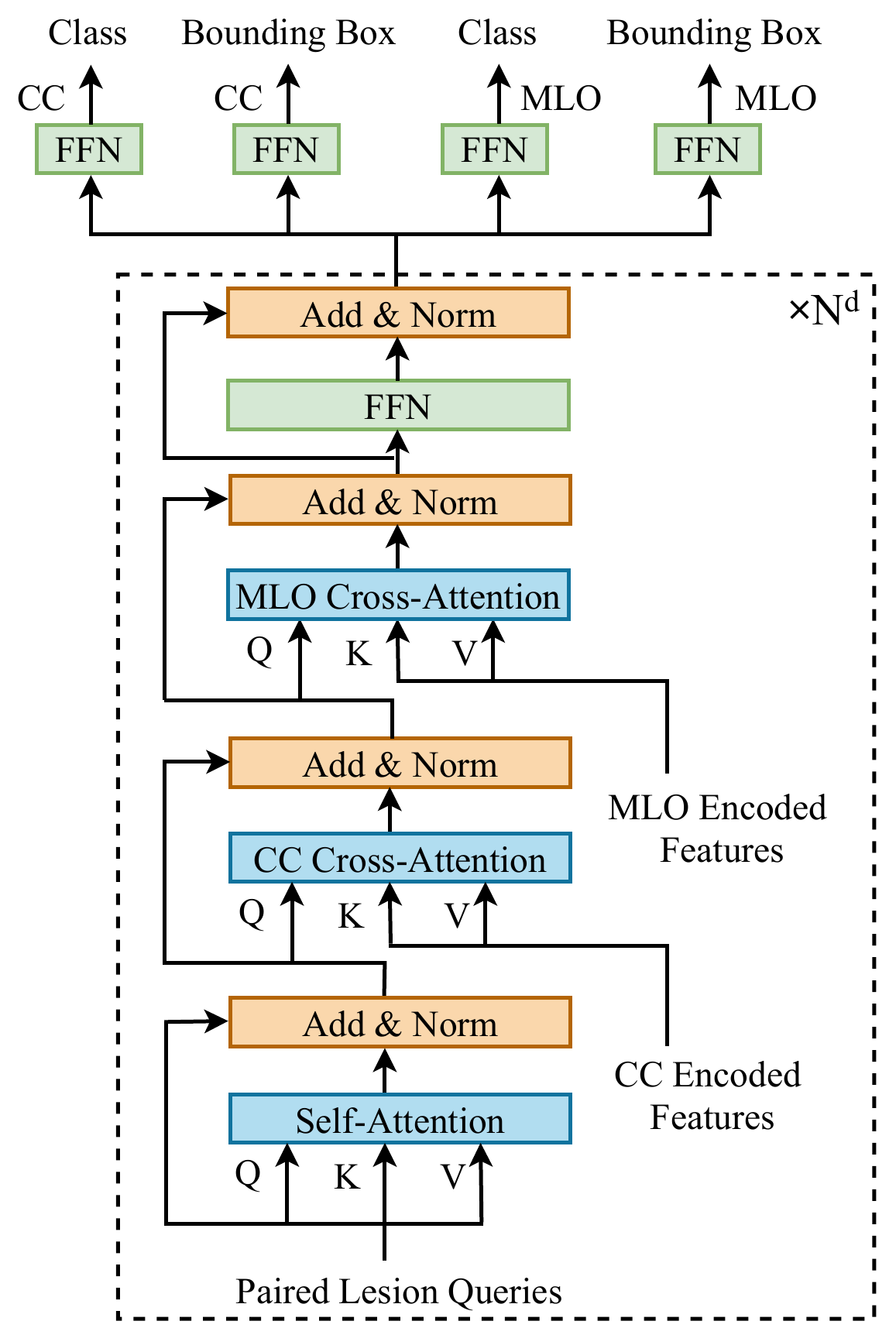}
\caption{Architecture of Decoder in Paired Lesion Query.}
\label{fig:decoder}
\end{figure}

\subsubsection{Paired Lesion Query}
We can also predict pairwise lesions with query mechanism directly. For this method, we use the same architecture of backbone and transformer encoder as VILD. In transformer decoder, as illustrated in Figure~\ref{fig:decoder}, we initialize a set of learnable paired lesion queries. At first paired lesion queries are passed through the multi-head self-attention layer. Then they will interact with image features for CC and MLO views outputted by transformer encoder sequentially to extract pairwise lesion information. Finally, a FFN layer is utilized to enhance the representative ability. Above layers can be stacked several times, and we denote the number of stacked layers as $\text{N}^\text{d}$. At the top of decoder, the classes and bounding boxes of MLO and CC views are directly predicted through several FFN layers for each query.

The output of decoder can be reformulated as a set of $N$ quads $\{\langle \hat{b}_i^c, \hat{b}_i^m, \hat{c}_i^c, \hat{c}_i^m \rangle \}_{i=1}^N$, where $\hat{b}_i^t \in \mathbb{R}^4$ and $\hat{c}_i^t \in \mathbb{R}^1$ are the predicted bounding box and classification score. Here $t \in \{\text{CC}, \text{MLO}\}$ denotes \text{CC} view or \text{MLO} view. The set of $N$ quads is similar with the extracted lesion pairs in lesion linker, thus we can also adopt a similar one-to-one matching assignment between the ground truth lesion pairs and the predicted lesion pairs. The cost function can be expressed as:
\begin{equation}
\begin{aligned}
\label{cost function}
     \mathcal{L}_{\text{match}}(y_i, \hat{y}_{\pi(i)}) 
     = & \mathbf{1}_{\{a_i \neq 0 \}} \cdot \text{max}\{ \mathcal{L}_{\text{CC}}(i,\pi(i)), \\
     & \mathcal{L}_{\text{MLO}}(i,\pi(i))\},
\end{aligned}
\end{equation}
where $\mathcal{L}_{t}$ is the same cost function as in~\cite{carion2020end} which contains classification cost and regression cost. $t \in \{\text{CC}, \text{MLO}\}$ denotes CC view or MLO view. The notation in Equation~\ref{cost function} is the same as Equation 17 in the main text. Instead of calculating the average of $\mathcal{L}_{\text{CC}}$ and $\mathcal{L}_{\text{MLO}}$, we adopt the larger one of the two costs. Since if one cost is significantly lower than the other, averaging them will let matching process be biased to the lower one. The final loss function can be written as follows:
\begin{eqnarray}
    \mathcal{L} = \mathcal{L}_\text{D}^\text{CC} + \mathcal{L}_\text{D}^\text{MLO},
\end{eqnarray}
where $\mathcal{L}_\text{D}^t$ is the loss function in DETR, and $t \in \{\text{CC}, \text{MLO}\}$ denotes CC view or MLO view. 

\subsubsection{Implementation Details}
Experimental details of Pair Verification and Paired Lesion Query are almost the same as {\net} in the main paper. We will elaborate on the different parts in the following.

\textbf{Pair Verification.}
We set $\alpha=0.75$ and $\gamma=2.0$ for the focal loss $\mathcal{L}_{\text{V}}$.

\textbf{Paired Lesion Query.}
The number of layers in decoder $\text{N}^\text{d}$ is set to 3 by default.

\subsection{Visualization and Analysis}

\begin{figure}[t]
\centering
\includegraphics[width=0.5\linewidth]{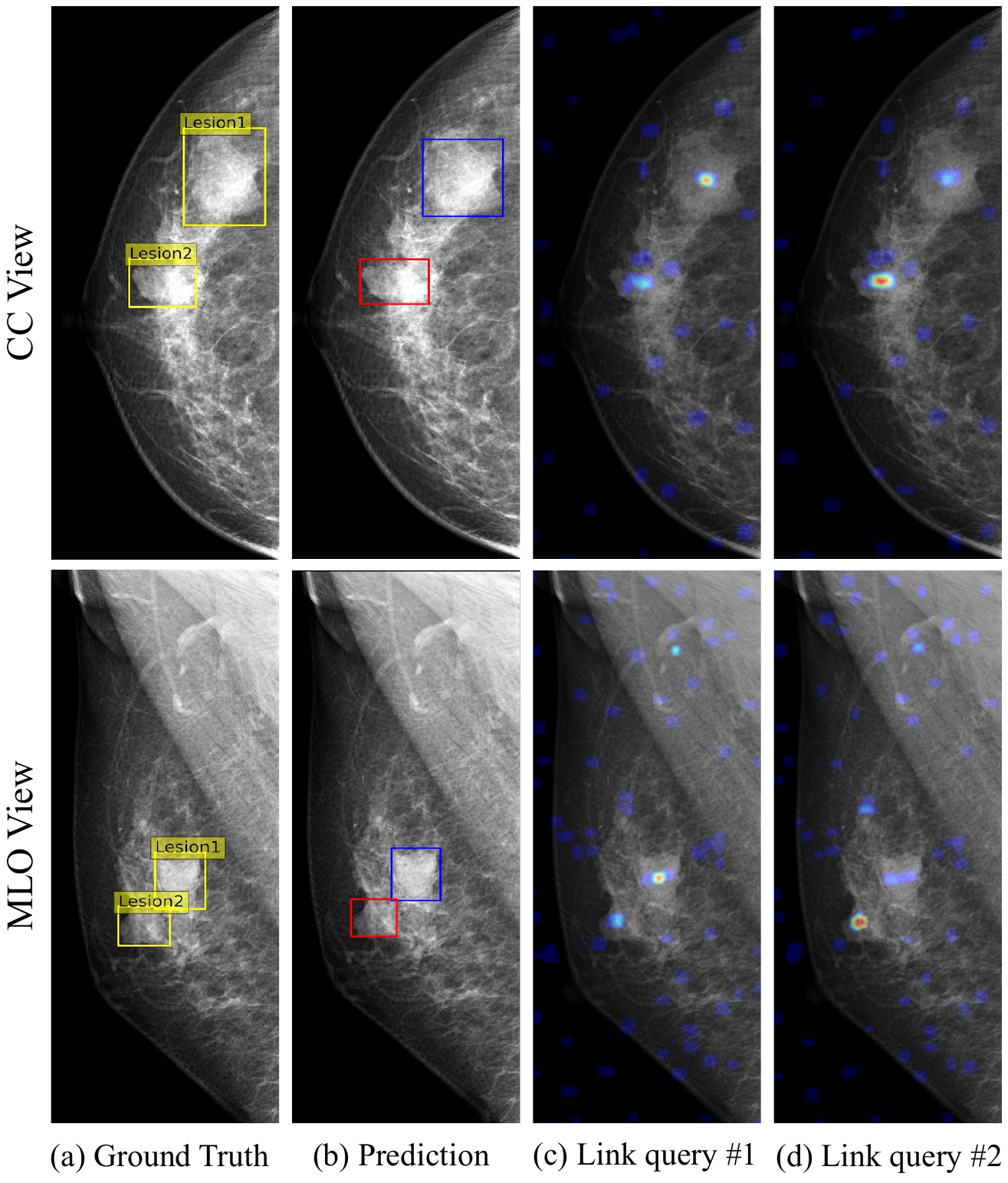}
\caption{\textbf{Case visualization.} We mark a lesion pair using the same id for ground truth and the same color for predictions. Column (c - d) visualize the attention weights between the two link queries that generate pair outputs in column (b) and the detection embeddings outputted by object queries of the two views. Red corresponds to larger weights.}
\label{fig:atten_map}
\end{figure}

Figure~\ref{fig:atten_map} illustrates an example of the detection and linking results of our proposed {\net}. As shown in column (a - b), the two predicted lesion pairs are consistent with the ground truth.
Column (c) and (d) present the attention weights between link queries and the detection embeddings outputted by object queries. 
For each link query, the highest responses are obtained from one lesion pair from the two views. This phenomenon tallies with our expectation that the lesion linker can learn the pairwise correspondence across views, which performs as the guidance for lesion detection.

\subsection{More Explanations about Dustbin Embedding}

\begin{figure}
\centering
\includegraphics[width=0.85\linewidth]{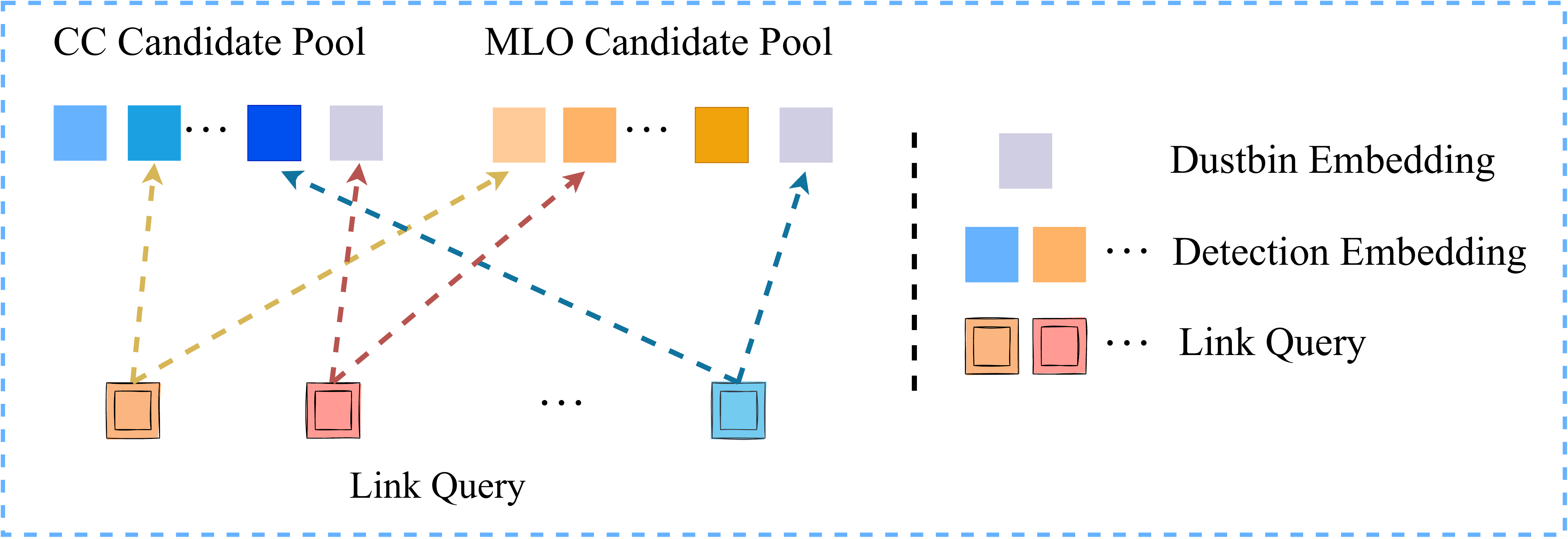}
\caption{The mechanism of dustbin embedding.}
\label{fig:dustbin}
\end{figure}

In Lesion Linker, we have firstly constructed the candidate pools of detection embeddings for MLO and CC views. Then, the link queries are responsible for cross-checking the suspicious detections and linking the same lesions in the two candidate pools. 
To handle the lesions which are visible only in one view, the mechanism of dustbin embedding is introduced.  The dustbin embedding is concatenated with the detection embeddings to endow Lesion Linker the ability to make predictions only in one view, which means the detection embedding is associated with the dustbin embedding by link query. The mechanism is illustrated in Figure~\ref{fig:dustbin}. Benefiting from the design of dustbin embedding, Lesion Linker can deal with different situations flexibly.

\subsection{Error Analysis}

\begin{wrapfigure}[19]{r}{0.5\textwidth}
\centering
\vspace{-20pt}
\includegraphics[width=1.0\linewidth]{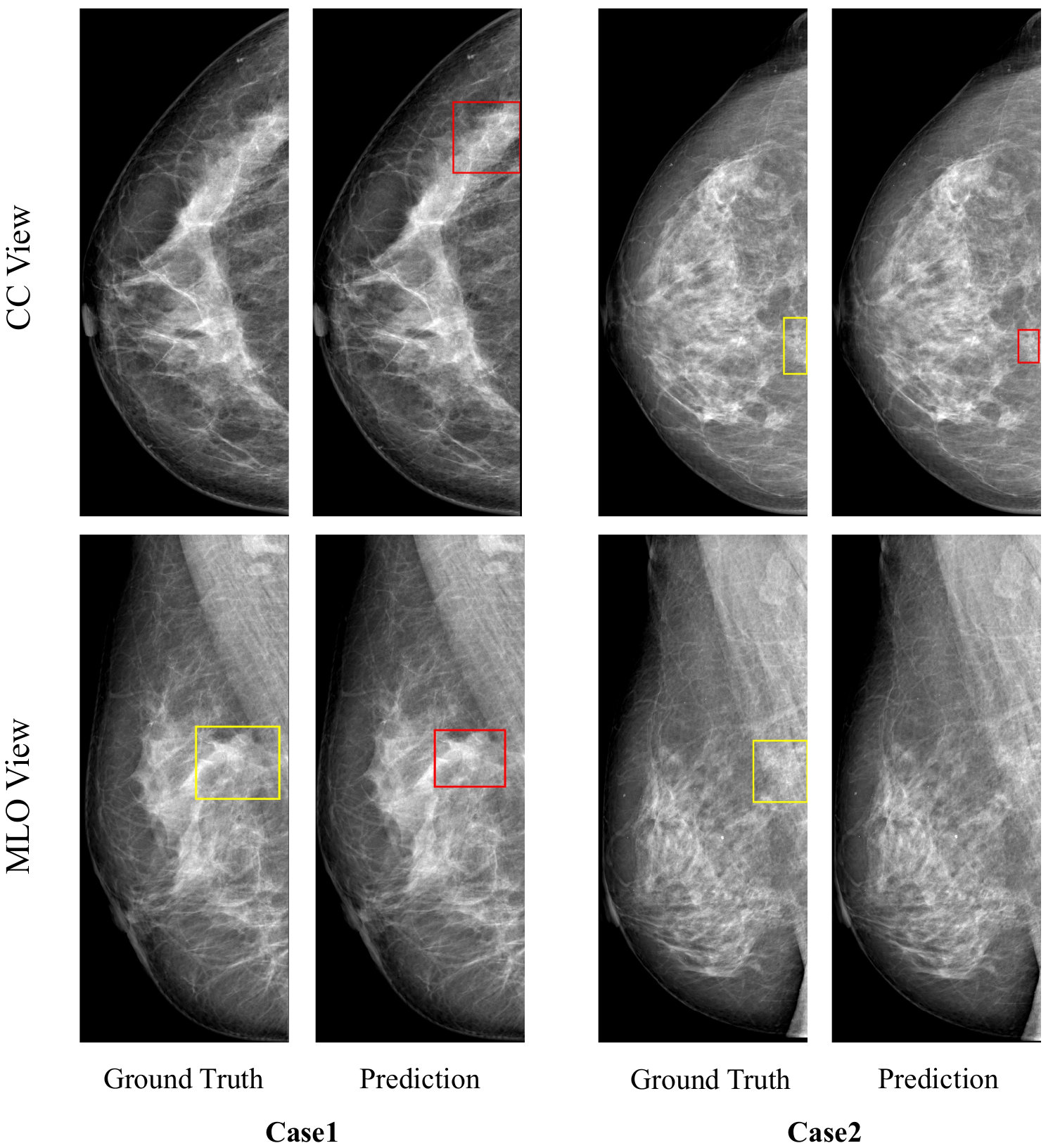}
\vspace{-20pt}
\caption{\textbf{Visualization of the cases with errors.}}
\label{fig:error}
\end{wrapfigure}

To analyze the limitations of the proposed approach, we have visualized some typical cases with obvious errors made by {\net}. As shown in the figure~\ref{fig:error}, there are mainly two types of mistakes: 1). the ground truth lesion could only be visible in MLO view (or CC view), while a FP box was detected in CC view (or MLO view). The FP box was associated with the TP box wrongly by Lesion Linker; 2). the model successfully discovered the ground truth lesion in one view while failed in the other view.

The main reason for these errors is that the ipsilateral information from mammogram is not sufficient for both deep learning models and radiologists to make an accurate diagnosis. One of the possible solutions is to leverage the bilateral (the same view of left and right breasts) information to further improve the detection performance, which could be our future work.

\clearpage
%
%
\bibliographystyle{splncs04}
\bibliography{egbib}
\end{document}